\documentclass[conference]{IEEEtran}
\IEEEoverridecommandlockouts
\usepackage{cite}
\usepackage{amsmath,amssymb,amsfonts}
\usepackage{algorithmic}
\usepackage{graphicx}
\usepackage{textcomp}
\usepackage{xcolor}
\usepackage{bm}
\usepackage{subcaption}
\usepackage{multirow}
\def\BibTeX{{\rm B\kern-.05em{\sc i\kern-.025em b}\kern-.08em
    T\kern-.1667em\lower.7ex\hbox{E}\kern-.125emX}}
\begin{document}

\title{Efficient Normalized Conformal Prediction and Uncertainty Quantification for Anti-Cancer Drug Sensitivity Prediction with Deep Regression Forests\\
\thanks{This work was supported in part by the National Science Foundation [2007903, 2007418] and NCI/Leidos under contract 22X049. The content is solely
the responsibility of the authors and does not necessarily represent the official
views of National Science Foundation or National Cancer Institute.}
}

\author{\IEEEauthorblockN{Daniel Nolte}
\IEEEauthorblockA{\textit{Dept. Electrical and Computer Engineering} \\
\textit{Texas Tech University}\\
Lubbock, USA \\
0000-0001-9631-8774}
\and
\IEEEauthorblockN{Souparno Ghosh}
\IEEEauthorblockA{\textit{Department of Statistics} \\
\textit{University of Nebraska - Lincoln}\\
Lincoln, USA \\
sghosh5@unl.edu}
\and
\IEEEauthorblockN{Ranadip Pal}
\IEEEauthorblockA{\textit{Dept. Electrical and Computer Engineering} \\
\textit{Texas Tech University}\\
Lubbock, USA \\
ranadip.pal@ttu.edu}
}

\maketitle

\begin{abstract}
Deep learning models are being adopted and applied on various critical decision-making tasks, yet they are trained to provide point predictions without providing degrees of confidence. The trustworthiness of deep learning models can be increased if paired with uncertainty estimations. Conformal Prediction has emerged as a promising method to pair machine learning models with prediction intervals, allowing for a view of the model's uncertainty. However, popular uncertainty estimation methods for conformal prediction fail to provide heteroskedastic intervals that are equally accurate for all samples. In this paper, we propose a method to estimate the uncertainty of each sample by calculating the variance obtained from a Deep Regression Forest. We show that the deep regression forest variance improves the efficiency and coverage of normalized inductive conformal prediction on a drug response prediction task.
\end{abstract}


\section{Introduction}\label{sec1}
In the field of machine learning (ML), a significant part of research focuses on the generalization error of the trained models, largely in terms of accuracy for classification or mean squared error for regression. ML is being used at an increasing rate for life-dependent decision-making across the healthcare field for example on tasks such as diagnosis or treatment recommendations \cite{sahni2023artificial,plana2022randomized,sun2022machine}.  In these extreme-stakes decision-making cases, minimizing error isn’t always enough, especially with deep networks that act like black boxes \cite{volovici2022steps}. With the increase in real-world ML medical applications, ML model predictions should be paired with a degree of confidence to enable insight into the model’s predictive uncertainty for each sample. 

One method that has recently seen an increase in popularity for conveying the predictive confidence of ML models is Conformal Prediction (CP) \cite{vovk2005algorithmic,papadopoulos2002inductive}. CP is a powerful method that guarantees marginally valid prediction intervals under the assumption that the data is exchangeable, or i.i.d.  CP can be performed with any ML model and even on pre-trained models without the need for further training, making it a powerful and flexible method. However, traditional CP provides constant intervals for all samples which is often not the best representation of uncertainty in prediction. Improvements to CP have been explored to introduce local adaptivity, the ability to have wider intervals for harder-to-predict samples, and conversely, smaller intervals for less uncertain samples. One popular method due to its simplicity is normalized conformal prediction, which normalizes the nonconformity scores based on an estimated degree of difficulty for each sample. Common methods for estimating the difficulty of each sample include measuring the variance of ensemble predictions, training an additional model to predict the error of the sample\cite{hernandez2022conformal}, and measuring the variance using Monte-Carlo dropout (MCD) with a neural network\cite{cortes2019reliable}. 

There has been an increasing amount of work in the field of CP recently, specifically with inductive conformal prediction (ICP) due to its flexibility and practicality. Some work focuses on improving ICP through adaptivity and the ability to handle distribution shift \cite{tibshirani2019conformal,barber2023conformal}, while others focus more on applications of CP such as in computational drug discovery \cite{cortes2019concepts} or medical imaging \cite{lu2022fair} for example. ICP has been successfully applied to numerous problems, specifically in drug design \cite{hernandez2022conformal,cortes2019reliable}. For instance, \cite{hernandez2022conformal} investigated using normalized ICP on the prediction of anti-cancer inhibitory activity of molecules and showed that ICP can be used as a tool to guide phenotypic drug design. For normalization, they trained two models, one as the point predictor and another to estimate the prediction error or sample uncertainty. Instead of training an additional model to predict the sample errors, \cite{cortes2019reliable} showed that test time MCD to estimate the sample uncertainties is an effective normalization and prediction error estimation method on 24 bioactivity data sets. 

In this paper, we instigate the use of Deep Regression Forests (DRF) for normalizing conformal predictions to provide more efficient instance-wise uncertainty estimations on an anticancer drug sensitivity prediction task. DRFs pair deep neural networks for feature extraction with probabilistic trees for ensemble predictions \cite{shen2018deep}.
We show that traditional normalization methods, while valid, fail to provide accurate conditional coverage across different target ranges. Additionally, we show that using the DRF variance for normalization leads to more efficient and conditionally valid Prediction Intervals (PI) which could thereby increase medical practitioner's trust in the model predictions.

\section{Related Work}\label{sec2}

\subsection{Adaptive Conformal Prediction}\label{subsec2_1}
There has been an increasing amount of work in the field of CP recently, specifically with inductive conformal prediction (ICP) due to its flexibility and practicality. Some work focuses on improving ICP through adaptivity and the ability to handle distribution shift \cite{tibshirani2019conformal,barber2023conformal}, while others focus more on applications of CP such as in computational drug discovery \cite{cortes2019concepts} or medical imaging \cite{lu2022fair} for example.

In addition to normalized CP for adaptivity, other methods include conformalized quantile regression \cite{romano2019conformalized} and Mondrian CP \cite{bostrom2020mondrian}. Conformalized quantile regression trains two models to predict the upper and lower bounds of the interval instead of training a point predictor. To avoid re-training a pre-trained model or having to train multiple models, there are extensions of CP, namely class-conditional and group-balanced CP for classification and regression tasks respectively \cite{vovk2005algorithmic, angelopoulos2021gentle}. These methods, also known as Mondrian CP, run the CP procedure for each sub-group/class separately, allowing for marginal validity for each group/class. Mondrian CP divides the calibration dataset into subgroups according to a Mondrian taxonomy, which can be based on the properties of the samples, such as class or group, or can be based on groupings of the uncertainties of the samples as in \cite{bostrom2020mondrian}. 

ICP has been successfully applied to numerous problems, specifically in drug design, recent papers have investigated the applicability of normalized ICP \cite{hernandez2022conformal,cortes2019reliable}. \cite{hernandez2022conformal} investigated using normalized ICP on NCI-60 and showed that improved predictive performance can be achieved when using ICP. For normalization, they trained two models, one as the point predictor and another to estimate the prediction error or sample uncertainty. Instead of training an additional model to predict the sample errors, \cite{cortes2019reliable} shows that test time MCD to estimate the sample uncertainties for normalization is an effective method on 24 ChEMBL data sets. In this paper we compare the performance of normalized ICP using a residual estimator, MCD, and DRF for the sample uncertainty predictions. 

\subsection{Deep Regression Forests}\label{subsec2_2}
The pairing of neural networks and decision trees has been sparking interest recently with many pairings trying to bring out the best of both worlds since NNs and DTs each have their own limitations and strengths \cite{li2022survey}. DRFs are one pairing that has shown state-of-the-art results on image tasks, such as age estimation \cite{shen2018deep}, head pose estimation \cite{ma2021learning}, and crowd counting \cite{ji2020hybrid}. Deep Neural Decision Forests (DNDF) \cite{kontschieder2015deep} were initially introduced as a state-of-the-art classifier on ImageNet, and not long after DRFs were adapted from DNDFs for use on regression problems. DRFs were further modified to fit ordinal regression problems \cite{zhu2021convolutional,lei2022meta}, and improved upon for early detection and classification of lung nodules from chest computed tomography scans \cite{lei2020meta}. To increase the fairness of DRFs, \cite{pan2020self} proposed Self-Paced forests with consideration of underrepresented examples (SPUDRFs). To estimate underrepresented examples, they propose using a measure of the entropy of the predictive output of the trees. However, the entropy approximation is ill-defined for small variances, which could easily be the case if the regression target range is small or normalized. In this paper, we instead aggregate the variance of a sample from a tree as a mixture distribution and directly use the variance as an estimation of sample uncertainty. Furthermore, we apply that uncertainty to the normalization of ICP intervals rather than applying it for resampling to accommodate underrepresented samples.

\section{Background/Methods}\label{sec3}

To formalize CP, consider a regression set, $\mathcal{Z}$, containing i.i.d. samples $z_i=(x_i \in \mathbb{R}^d, y_i \in \mathbb{R})$, where each $z_i$ consists of a d-dimensional vector of features, $x_i$, and a response variable, $y_i$. With no assumptions on the distribution of the data or the underlying predictive model, $f(x_i)$, the goal of CP is to construct a prediction interval $C(x_i)$ for a new feature vector, $x_{n+1}$, given a misscoverage rate, $\alpha$, or Confidence Level (CL), $1-\alpha$. CP provides the guarantee that the unobserved response variable $y_{n+1}$ is covered by the interval with probability $1-\alpha$:
\begin{equation}\label{eq1}
\mathbb{P}[y_{n+1} \in  C(x_{n+1})] \geq 1-\alpha
\end{equation}
CP is a flexible method as it makes minimal assumptions about the data, except for exchangeability, and can be applied on nearly any pre-trained point-predictor in which we can define a non-conformality score for each sample. The non-conformality score quantifies the disagreement between model prediction and response variable and is typically the mean absolute error for regression problems:
\begin{equation}\label{eq2}
a(x_i,y_i) = | y_i - f(x_i) |
\end{equation}

Conformal Prediction CP was initially designed to be re-trained for each new sample which made the method computationally expensive and unreasonable. Inductive conformal prediction (ICP), also known as split-conformal prediction, was later introduced to mitigate this issue by using only one trained model at the expense of using part of the training data for CP calibration.

\subsection{Inductive Conformal Prediction}\label{subsec3_1}

For ICP, the dataset, $\mathcal{Z}$, must be split into proper training and calibration datasets, $\mathcal{Z}_t$ and $\mathcal{Z}_c$ respectively. Once the predictive model, $f$, is trained on $\mathcal{Z}_t$, ICP uses the model to obtain non-conformality scores for each sample in $\mathcal{Z}_c$. This produces a list of scores, $A = a_1,...,a_m$ where $m = |\mathcal{Z}_c|$. Next, $\hat{q}$ is taken as the $(1-\alpha)$ quantile of $A$. Then for future samples, ICP returns the prediction intervals:
\begin{equation}\label{eq3}
C(x_{n+1}) = [f(x_{n+1})-\hat{q}, f(x_{n+1})+\hat{q}].
\end{equation}
As seen in equation (\ref{eq3}), traditional ICP returns a constant interval for each sample, specifically with width $2*\hat{q}$. ICP has guarantees of valid marginal coverage satisfying equation (\ref{eq1}).
However, since the intervals are constant for all samples, the intervals fail to provide adaptivity. Adaptivity can be considered through the lens of conditional coverage \cite{vovk2012conditional,angelopoulos2021gentle}:
\begin{equation}\label{eq4}
\mathbb{P}[y_{n+1} \in  C(x_{n+1}) |  x_{n+1}] \geq 1-\alpha
\end{equation}
Conditional coverage is much stronger than marginal coverage as now it tries to return valid PIs for each input $x_i$. Conditional coverage isn’t a guarantee that comes with any general ICP method, in fact, in the most general sense it is impossible to achieve with finite samples \cite{vovk2012conditional,lei2014distribution}. Thus, conditional coverage is an important proxy to evaluate the adaptivity of the different methods. 

In addition to coverage validity, efficiency is another metric often used to measure how effective the PIs are. In regression tasks, the efficiency is often the average interval width where smaller intervals are preferred while maintaining valid $(1-\alpha)$ PIs. 
Validity exceeds efficiency in terms of priority as it is necessary to form valid PIs, while efficiency helps measure how effective the different methods are. 

A simple method to add adaptivity into ICP, while often improving efficiency, is through normalized, or locally adaptive, conformal prediction. With normalized ICP, the non-conformality scores are normalized with sample-level predictions of uncertainty, $\sigma_i$:
\begin{equation}\label{eq5}
a(x_i,y_i) = \dfrac{| y_i - f(x_i) |}{\sigma_i+\beta} 
\end{equation}
with $\beta$ as a smoothing parameter. Then the prediction intervals for new samples are computed as:
\begin{equation}\label{eq6}
C(x_{n+1}) = [f(x_{n+1})-\hat{q}(\sigma_{n+1}+\beta), f(x_{n+1})+\hat{q}(\sigma_{n+1}+\beta)].
\end{equation}
with $\hat{q}$ calculated the same as above.

The measure of uncertainty, $\sigma_i$, can be approximated in multiple different ways. Some popular methods include approximating $\sigma_i$ as the variance of ensemble predictions, as the variance of multiple Monte Carlo Dropout (MCD) model predictions, or fitting a second model to estimate the residuals of the first model given the input features. 

In the rest of this paper, we focus on the adaptivity and efficiency of different normalization methods for normalized ICP. We investigate the effects of using MCD, secondary residual fitted model, and DRF leaf variance as the normalization uncertainty measure of each sample. For a more in-depth study on conformal prediction, we refer the readers to the recent reviews \cite{angelopoulos2021gentle,fontana2023conformal,lei2018distribution}.

\subsection{Deep Regression Forests for Uncertainty Estimation}\label{subsec3_2}
DRFs combine the rich representational feature extraction of deep neural networks with ensemble predictions \cite{shen2018deep,nolte2023federated}. To achieve this, the trees in the forest are probabilistic, rather than deterministic as in the case of random forests. The extracted rich representational output features of a neural network are transformed into probabilities by a sigmoid function and used as the routing probability of the tree nodes. Formally, DRFs consist of multiple deep regression trees. Each tree is a binary tree with split nodes $\mathcal{N}$ and leaf nodes $\mathcal{L}$. Each split node, $n\in\mathcal{N}$, consists of a split function that guides the input $x_i$ to the right or left subtree:
\begin{equation}\label{eq7}
s_n(x_i;\Theta) = \sigma(f_{\varphi(n)}(x_i;\Theta))
\end{equation}
where $\sigma(\cdot)$ is the sigmoid function and $\varphi(n)$ is an indexing function to map the neural network outputs, $f(x_i;\Theta)$, to split nodes $n$, with neural network parameters $\Theta$ \cite{shen2018deep}. 

Each leaf node consists of a probability density distribution, typically modeled through a Gaussian distribution with mean $\mu_l$ and variance $\sigma_l^2$, over the target space. The probability of reaching a leaf node is the product of all preceding parent nodes' routing probabilities in the tree:
\begin{equation}\label{eq8}
P(\ell|x;\Theta) = \prod_{n\in\mathcal{N}} s_n(x;\Theta)^{\textbf{1}(\ell\in\mathcal{L}_{n_l})}(1-s_n(x;\Theta))^{\textbf{1}(\ell\in\mathcal{L}_{n_r})}
\end{equation}
with $\textbf{1}(\cdot)$ denoting the indicator function, $\mathcal{L}_{nl}$ and $\mathcal{L}_{nr}$ corresponding to the sets of leaf nodes owned by subtrees of parent node n, and with $n_l$ and $n_r$ as the left and right children of node n. 
The predictions of the trees are the routing probability-weighted sum of the expected values of the leaf node distributions and can be viewed as a mixture distribution with mixing coefficients $P(\ell|x;\Theta)$. With the leaf node distribution parameters for a tree denoted by $\pi$ and the overall forest leaf node distribution parameters denoted by $\Pi$, the point prediction from a tree is given by:
\begin{equation}\label{eq9}
p_{\mathcal{T}}(y_i|x_i;\Theta,\pi) = \sum_{\ell\in\mathcal{L}} P(\ell|x;\Theta)\mu_\ell
\end{equation}
Thus, with $K$ trees in the forest, the prediction from the entire forest is the average across all tree predictions:
\begin{equation}\label{eq10}
p_{\mathcal{F}}(y_i|x_i;\Theta,\Pi) = \frac{1}{K}\sum_{k=1}^K p_{\mathcal{T}_k}(y_i|x_i;\Theta,\pi_k) 
\end{equation}

DRFs thus must learn both the network parameters as well as the leaf node distribution parameters. The networks used for feature extraction are typically CNNs, as in \cite{shen2018deep,ma2021learning,ji2020hybrid}, but any back-propagation compatible model can be used. Training consists of 2 steps, one to update the neural network parameters through typical backpropagation while the leaf parameters are frozen, and another to update the leaf node parameters through variational bounding while the network parameters are frozen. Interestingly, in addition to aggregating the sample probability-weighted mixture distribution mean for a point prediction for each sample, we can also aggregate the mixture distribution variance for an estimation of the uncertainty for each sample. Formally, the variance of a sample from a deep regression tree can be calculated as:
\begin{equation}\label{eq11}
\begin{aligned}
\sigma_{\mathcal{T}}(y_i|x_i;\Theta,\pi)^2 = \\
\sum_{\ell\in\mathcal{L}} P(\ell|x;\Theta)\sigma_\ell^2+P(\ell|x;\Theta)\mu_\ell^2-(P(\ell|x;\Theta)\mu_\ell)^2
\end{aligned}
\end{equation}
and as above, we can calculate the forest variance as the average across all trees.
\begin{equation}\label{eq12}
\sigma_{\mathcal{F}}(y_i|x_i;\Theta,\Pi)^2 = \frac{1}{K}\sum_{k=1}^K \sigma_{\mathcal{T}}(y_i|x_i;\Theta,\pi_k)^2
\end{equation}

Additionally, as in typical ensemble uncertainty predictions like MCD, we can also calculate the variance of the tree predictions across the forest to act as another measure of uncertainty:
\begin{equation}\label{eq13}
\sigma_{Ensemble}^2 = Var[p_{\mathcal{T}_1}(y_i|x_i;\Theta,\pi_1),...,p_{\mathcal{T}_K}(y_i|x_i;\Theta,\pi_K)]
\end{equation}


\begin{figure*}[ht]
\begin{center}
  \begin{subfigure}{0.29\textwidth}
    \includegraphics[width=\linewidth]{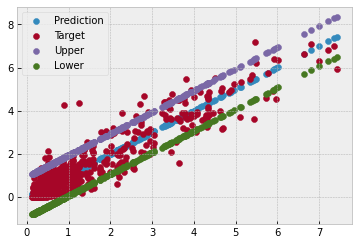}
    \caption{ANN CP}
  \end{subfigure}%
  \hspace*{\fill}   
  \begin{subfigure}{0.29\textwidth}
    \includegraphics[width=\linewidth]{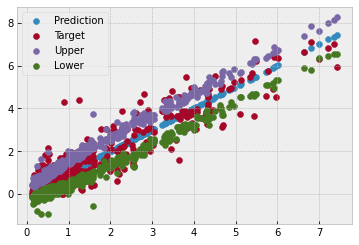}
    \caption{ANN RF} 
  \end{subfigure}%
  \hspace*{\fill}   
  \begin{subfigure}{0.29\textwidth}
    \includegraphics[width=\linewidth]{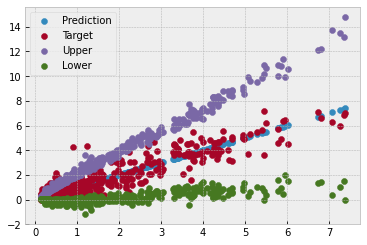}
    \caption{ANN MCD}
  \end{subfigure}
  \hspace*{\fill}
  \begin{subfigure}{0.29\textwidth}
    \includegraphics[width=\linewidth]{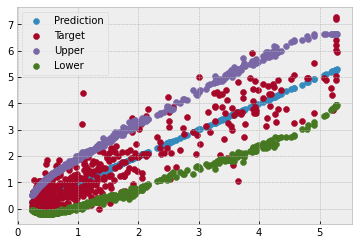}
    \caption{DRF STD}
  \end{subfigure}%
  \hspace*{\fill}   
  \begin{subfigure}{0.29\textwidth}
    \includegraphics[width=\linewidth]{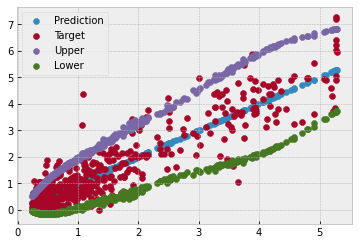}
    \caption{DRF STD with Ensemble STD}
  \end{subfigure}
  \hspace*{\fill}
  \caption{CCLE Prediction Intervals for the 5 competing methods with CL set to 90\% for 500 test samples}
  \label{fig:Intervals}
\end{center}
\end{figure*}

\section{Results and Discussion}\label{sec4}
\subsection{Dataset}
Since a large motivating factor in explainable ML can be contributed to healthcare applications, we chose to assess the performance of competing normalized ICP methods on a human anticancer drug efficacy prediction task. The cancer cell line encyclopedia (CCLE) contains a detailed genetic and pharmacologic characterization of a large panel of human cancer cell lines \cite{barretina2012cancer,cancer2015pharmacogenomic,ghandi2019next,rahman2019evaluating}. CCLE contains the inhibitory responses, expressed in terms of IC50, of 24 anticancer drugs screened on 504 different cancer cell lines with known genomic information. The data was obtained and pre-processed the same as in \cite{nolte2023federated}. The dataset contains two inputs (a) PaDEL descriptors \cite{yap2011padel} of the 24 anti-cancer molecules and (b) microarray gene expressions of the 504 cell lines before drug application. The response consists of the area under the curve (AUC) of the drug responses for a particular drug applied on a particular cell line. The preprocessed dataset contained 10,838 AUC measurements, with 1072 chemical descriptor features and 1101 gene expression features. To form the full feature space, the chemical descriptors and gene expressions were concatenated to create a full feature vector for each drug-cell pair.
\subsection{Competing Methods}\label{subsec4_1}
Several different methods for normalizing ICP exist. Popular methods include training an additional model to predict the residuals given the feature vector, using the variance of an ensemble of models, and using MCD to simulate ensemble predictions from a neural network to obtain the variance of predictions. We compare the normalization performance of different uncertainty predictions using a random forest (RF) residual model, MCD, and the variance obtained from DRFs. To test non-normalized ICP as well as MCD and RF uncertainties, we paired them with the predictions of an artificial neural network (ANN). Each ANN or DRF method uses the same trained model for each different ICP method. The 5 different competing methods are described in further detail below.

\noindent\textbf{ANN CP:} Traditional (non-normalized) ICP with ANN as the point predictor.

\noindent\textbf{ANN RF:} Normalized ICP with ANN as the point predictor and a Random Forest as the uncertainty prediction model.

\noindent\textbf{ANN MCD:} Normalized ICP with ANN as the point predictor and MCD for prediction uncertainty. 

\noindent\textbf{DRF STD:} Normalized ICP with DRF as the point predictor and DRF STD for the prediction uncertainty.

\noindent\textbf{DRF STD + Ensemble STD:} Normalized ICP with DRF as the point predictor and DRF STD + Ensemble STD for the prediction uncertainty.

\subsection{Model Training}\label{subsec4_2}
For proper evaluation of the methods, the CCLE dataset was split into $\mathcal{Z}_t$, $\mathcal{Z}_c$, and $\mathcal{Z}_{test}$ with an 80/10/10\% split. This was performed for 5 random partitions of the data and the results were averaged. The models were trained on $\mathcal{Z}_t$, the ICP was calibrated with $\mathcal{Z}_c$, and $\mathcal{Z}_{test}$ was held out for reporting the evaluation results. 

The ANN architecture was determined through a grid search over the number of layers, number of neurons in each layer, learning rate, and with and without batch normalization. The optimal model contained 7 layers with neuron sizes 1500, 1000, 600, 300, 100, 50, and 1, all without batch normalization. All ANN layers were followed by a dropout layer with a drop probability of 10\%. The DRF architecture was based on the optimized ANN architecture with the last 3 layers replaced by a forest of 15 trees, each with a depth of 7. The main difference between the architectures, besides the forest, was that the DRF benefited from the addition of batch normalization, and the output of the 4th layer was changed from 300 to 600 for generating the routing probabilities for the trees.

The models were trained using Python 3.8 and the PyTorch package. For optimization, the Adam optimizer was used and the learning rate was set to 1e-4 with a batch size of 256. Since $\mathcal{Z}_c$ was held out of training, it was used as a validation set for early stopping and learning rate scheduling. Once the model had 5 consecutive epochs without improvement, the learning rate was reduced by a factor of 10. Additionally, after 10 consecutive epochs without improvement, the training was halted and the best-performing model was saved. 

Once the models were trained, the saved model was used to calculate the non-conformality scores of each sample of $\mathcal{Z}_c$ using (\ref{eq2}) for traditional ICP and (\ref{eq5}) for normalized ICP. The scores were then sorted in ascending order and the $(1-\alpha)$ percentile of the non-conformality scores was taken as $\hat{q}$. Finally, the estimated intervals were calculated for the test samples in $\mathcal{Z}_{test}$ using (\ref{eq3}) for traditional ICP and (\ref{eq6}) for normalized ICP. The results of each method are averaged across all 5 random partitions. In our experiments, we vary the confidence level, $1-\alpha$, with values of 70\%, 80\%, and 90\% for all competing methods and set $\beta$ to 0. 

\begin{table*}[ht]
\centering
\caption{CCLE Conditional Coverage Results averaged across all 5 partitions with a varying CL}
\begin{tabular}{c|c|cccccc}\hline 
\multicolumn{1}{l}{} &
  \multicolumn{1}{l}{\textbf{}} &
  \textbf{ANN CP} &
  \textbf{ANN MCD} &
  \textbf{ANN RF} &
  \textbf{DRF STD} &
  \textbf{DRF STD + Ensemble STD} \\\hline  
  \multicolumn{1}{l}{\textbf{}}   & \bm{$R^{2}$}             & 0.85182 & 0.85175 & 0.85182 & 0.84416 & 0.84416 \\
\multicolumn{1}{l}{\textbf{}}   & \textbf{PCC of uncertainty and error}            & N/A      & 0.33682 & 0.30112 & \textbf{0.38339} & 0.37793  \\\hline
                                  & \textbf{Coverage Low ($y_i\leq 2$)}              & 0.77102 & 0.65618 & 0.66225 & 0.73872 & 0.72625 \\
                                & \textbf{Coverage Med($2< y_i \leq4$)}             & 0.48562 & 0.84534 & 0.57261 & 0.69401   & 0.70353 \\
                                & \textbf{Coverage High($y_i>4$)}           & 0.44949 & 0.92553 & 0.54003 & 0.67749  & 0.69864 \\
\multirow{-4}{*}{\textbf{70\%}} & \textbf{MAD Conditional Coverage} & 0.17864 & 0.13823 & 0.10837 & 0.02241 & \textbf{0.01038} \\\hline  
                                & \textbf{Coverage Low ($y_i\leq 2$)}              & 0.86164 & 0.77273 & 0.69613 & 0.84038 & 0.83063 \\
                                & \textbf{Coverage Med($2< y_i \leq4$)}           & 0.58458 & 0.89257 & 0.59629 & 0.76109  & 0.77271 \\
                                & \textbf{Coverage High($y_i>4$)}           & 0.55062 & 0.96019 & 0.55485 & 0.75715 & 0.78576 \\
\multirow{-4}{*}{\textbf{80\%}} & \textbf{MAD Conditional Coverage} & 0.17548 & 0.09334 & 0.18424 & 0.04071 & \textbf{0.02405} \\\hline  
                                & \textbf{Coverage Low ($y_i\leq 2$)}              & 0.95006 & 0.89819 & 0.73119 & 0.94042 & 0.93725 \\
                                & \textbf{Coverage Med($2< y_i \leq4$)}            & 0.75314 & 0.92513 & 0.63167 & 0.87264 & 0.87384 \\
                                & \textbf{Coverage High($y_i>4$)}           & 0.72259 & 0.97909 & 0.58924 & 0.84704 & 0.85286 \\
\multirow{-4}{*}{\textbf{90\%}} & \textbf{MAD Conditional Coverage} & 0.12478 & \textbf{0.03534} & 0.24930  & 0.04025 & \textbf{0.03684} \\\hline  
\end{tabular}
\label{table:condCoverage}
\end{table*}

\begin{table*}[!htp]
\centering
\caption{CCLE Normalized ICP coverage and efficiency averaged across 5 random partitions with a varying CL}
\begin{tabular}{c|c|cccccc}\hline 

\multicolumn{1}{l}{} &
  \multicolumn{1}{l}{\textbf{}} &
  \textbf{ANN MCD} &
  \textbf{ANN RF} &
  \textbf{DRF STD} &
  \textbf{DRF STD +   Ensemble STD} \\\hline
                                & \textbf{Coverage}     & 0.71070 & 0.63708 & 0.72601 & 0.72011 \\
\multirow{-2}{*}{\textbf{70\%}} & \textbf{Interval Width}  & 1.47967 & 0.91951 & \textbf{1.16082} & 1.16944 \\\hline
                                & \textbf{Coverage}      & 0.80885 & 0.66753 & 0.82029  & 0.81752 \\
\multirow{-2}{*}{\textbf{80\%}} & \textbf{Interval Width}  & 1.88772 & 0.97978 & \textbf{1.38464} & 1.40171 \\\hline
                                & \textbf{Coverage}      & 0.91014 & 0.70277 & 0.92122 & 0.91955 \\
\multirow{-2}{*}{\textbf{90\%}} & \textbf{Interval Width}  & 2.63836 & 1.06139  & \textbf{1.80832} & 1.82097\\\hline
\end{tabular}
\label{table:CpResults}
\end{table*}

\subsection{Metrics}\label{subsec4_3}
The coefficient of determination, or $R^2$, is reported to evaluate the performance of the point predictions. Moreover, to evaluate the performance of the uncertainty quantification, we report the Pearson correlation coefficient (PCC) of the predicted uncertainty and the model errors.  For evaluation of the prediction intervals, we report the coverage in terms of the percent of test sample target values that lie within the estimated prediction intervals. In addition to coverage, we also report the efficiency, or average interval width, across all test samples. Coverage paired with efficiency allows a view into how effective the normalization methods are, as smaller intervals with proper coverage often convey more information. Since conditional coverage can't be calculated directly, we proxy it by evaluating the marginal coverage for different target ranges while the overall marginal coverage is calculated across all target values.

\begin{figure}[ht]
\begin{center}
    \includegraphics[width=\linewidth]{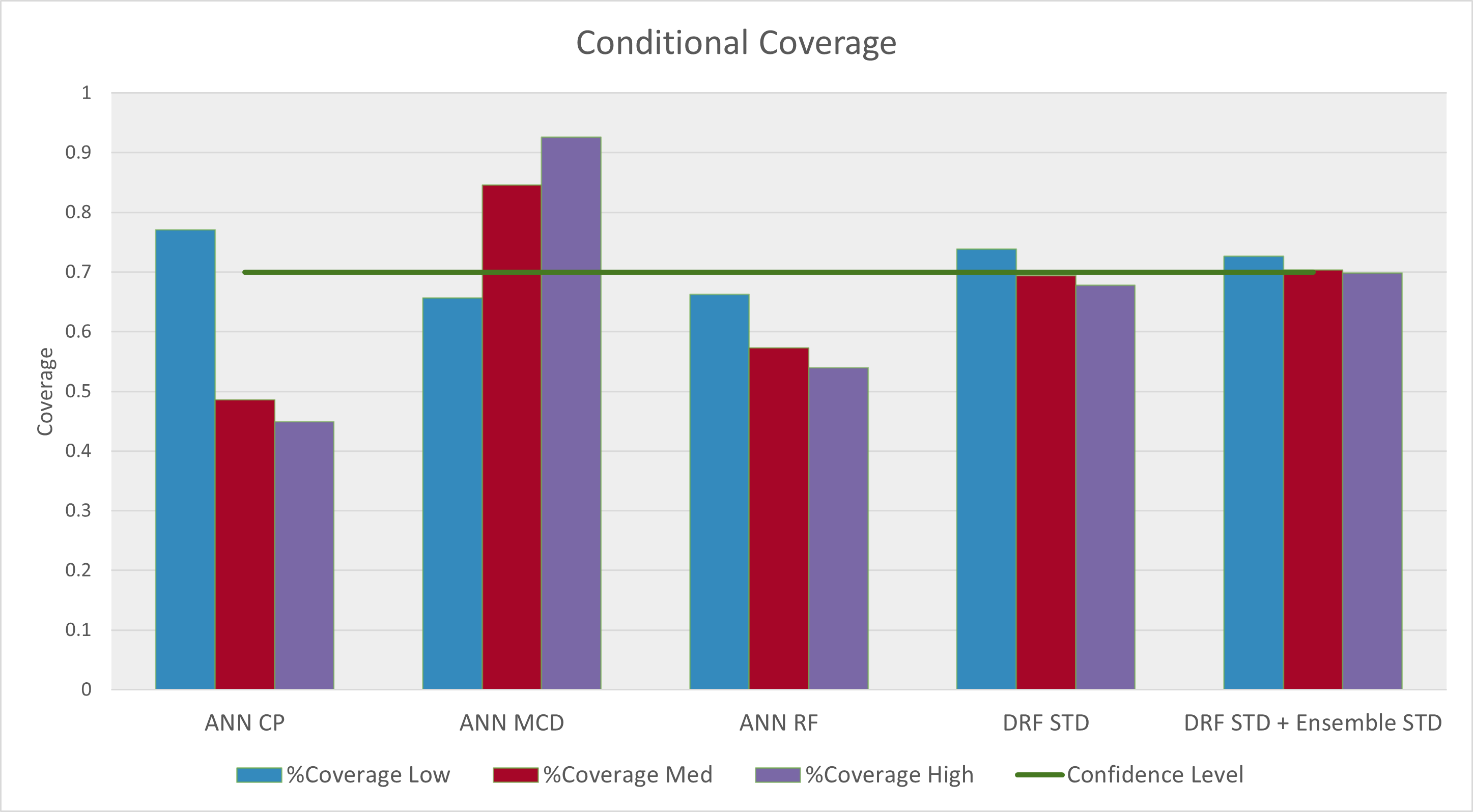}
  \caption{Conditional Coverage for each competing method with a CL of 70\%}
  \label{fig:CondCov}
\end{center}
\end{figure}

\subsection{CCLE ICP Results}\label{subsec4_4}
The average results across 5 random partitions for 3 different confidence levels are presented in table \ref{table:CpResults} and \ref{table:condCoverage}.  As shown, the ANN and DRF achieve comparable $R^2$. However, the PCCs show that the DRF standard deviations were more correlated with the model errors, indicating it's a more suitable method for sample-wise uncertainty quantification compared to MCD. All methods, except ANN RF, achieved accurate marginal coverage across different confidence levels due to the ICP procedure. The random forest model failed to successfully model the residual for the test set, indicating that more data is needed to properly train the model, which would decrease the predictive performance of the ANN given a finite sample space $\mathcal{Z}$. The MCD method provides correct coverage, but it returns large intervals, especially for larger target values. To display the heteroskedasticity and accuracy of the prediction intervals we plot the predicted intervals against the predicted values with the target values overlaid in Fig. \ref{fig:Intervals}. As demonstrated, the ANN MCD intervals are very wide for target values greater than 2, and the ANN CP method provides constant intervals that are too narrow for the same target range. Also, the ANN RF normalization method produces intervals that are too narrow and improperly fit for the test data. Looking at the DRF intervals, they are much smoother and more closely capture the heteroskedasticity of the model errors.

\subsection{Evaluating Adaptivity}\label{subsec4_5}
To further investigate the heteroskedasticity, we proxied conditional coverage by evaluating the coverage of 3 different target ranges separately. Specifically, since our target values range from 0-8, with most values between 0-4, we split the target ranges into Low, for $y_i \leq 2$, Medium (Med), for $2< y_i \leq4$, and High for $y_i>4$. The results for each range with each competing method are shown in table \ref{table:condCoverage} as well as the mean absolute deviation (MAD) of the conditional coverage from the specified confidence levels. As noted before, the traditional ANN CP method does not accurately capture the heterogeneity of errors across all target ranges. Since the intervals are constant, the method has to overestimate the intervals for the Low range while underestimating the intervals for the Med and High ranges. Conversely, the MCD method overestimated the Med and High ranges while underestimating the Low range. Finally, the DRF STD methods slightly overestimated the low range and underestimated the Med and High ranges. This is further shown through the spread of conditional coverage for CL 70\% in figure \ref{fig:CondCov}. Additionally, as indicated by the MAD conditional coverage, the DRF STD methods provide more accurate conditional coverage across different confidence levels. When comparing DRF STD with DRF STD + Ensemble STD, the ensemble STD can be viewed as a heteroskedastic smoothing parameter, $\beta$. When including the Ensemble STD, the intervals widen very slightly while allowing more accurate conditional coverage.

\section{Conclusion}\label{sec5}
In this paper, we have presented a method to aggregate the variance of DRF predictions to provide more accurate estimates of predictive uncertainty for normalized ICP. We applied our method to an anti-cancer drug-response prediction dataset, CCLE, and compared it to traditional normalization methods MCD and RF residual predictions. Results indicate the potential for using DRF based conformal predictions for providing uncertainty estimates for personalized medicine applications. We showcase the ability of DRF CP normalization to accurately capture the heteroskedasticity of the model predictions. Additionally, we demonstrate that the DRF normalization achieves better efficiency and accurate conditional coverage. Considering the success of this method, other mean-variance model predictors should be investigated for their ability to accurately capture the predictive uncertainty and improve upon ICP efficiency and conditional coverage.

\bibliographystyle{unsrt}
\bibliography{reference}

\end{document}